\documentclass[conference]{IEEEtran}

\makeatletter
\def\ps@IEEEtitlepagestyle{%
  \def\@oddfoot{\mycopyrightnotice}%
  \def\@evenfoot{}%
}
\def\mycopyrightnotice{%
  {\footnotesize 979-8-3195-3450-7/26/\$31.00~\copyright~2026 IEEE\hfill}
  \gdef\mycopyrightnotice{}
}

\usepackage{blindtext}
\usepackage{eso-pic}
\IEEEoverridecommandlockouts
\usepackage{cite}
\usepackage{amsmath,amssymb,amsfonts}
\usepackage{algorithmic}
\usepackage{graphicx}
\usepackage{textcomp}
\usepackage{xcolor}

\usepackage{hyperref}
\usepackage{booktabs}

\def\BibTeX{{\rm B\kern-.05em{\sc i\kern-.025em b}\kern-.08em
    T\kern-.1667em\lower.7ex\hbox{E}\kern-.125emX}}
    
\usepackage{eso-pic}

\usepackage{fancyhdr}

\newcommand{\arxivheader}{%
  \small Accepted for publication in the 2026 International Conference on Electrical and Computer Engineering Researches (ICECER 2026)
}

\newcommand{\arxivcopyright}{%
  \vbox{%
    \hrule height 0.4pt \vskip 5pt
    \small \noindent © 2026 IEEE. Personal use of this material is permitted. Permission from IEEE must be obtained for all other uses, in any current or future media, including reprinting/republishing this material for advertising or promotional purposes, creating new collective works, for resale or redistribution to servers or lists, or reuse of any copyrighted component of this work in other works.
  }%
}

\fancypagestyle{arxivstyle}{
  \fancyhf{}
  
  \fancyhead[C]{\arxivheader}   
  \fancyfoot[C]{\arxivcopyright} 
}

\begin{document}
\title{\vspace*{1cm} Copyright Leakage Detection in AI Generated Videos
}


\author{
\IEEEauthorblockN{
Shaif Chowdhury,
Moshiur Rahman Tonmoy
}
\IEEEauthorblockA{
\textit{Department of Electrical Engineering and Computer Science, Texas A\&M University--Kingsville}\\
Kingsville, TX 78363, USA\\
shaif.chowdhury@tamuk.edu\\
moshiur\_rahman.tonmoy@students.tamuk.edu
}
}

\maketitle
\thispagestyle{arxivstyle}

\begin{abstract}

Video generation models can create realistic-looking video footage from images and text prompts. Some of the recent advances have raised concerns about the reproduction of copyrighted visual content.
However, copyright leakage detection in generated videos remains largely unexplored. 

In this paper, we introduce a new benchmark data set for copyright leakage detection, comprising 50 copyrighted concepts, each with 5 images. For each concept, we have 5 videos created using the image+prompt video generation model and the prompt-only video generation model. Following that, we develop an AI toolset based on a contrastive retrieval framework that formulates copyright leakage detection as a video-to-image retrieval problem by matching aggregated video frame embeddings against a database of copyrighted image embeddings. Our method uses a single similarity threshold and retrieves the most similar copyrighted reference as supporting evidence. 

We conducted experiments showing strong retrieval and detection performance, achieving 99.6\% detection accuracy with only a single false negative. These results show the effectiveness of the retrieval-based method for scalable and explainable copyright leakage detection in AI-generated videos. Data can be accessed publicly from \url{https://zenodo.org/records/21386117} and the source code for this project is publicly available at
\url{https://github.com/Shaif95/video-cp-Leakage}. 

\end{abstract}

\begin{IEEEkeywords}
video generation, copyright leakage, video understanding, contrastive learning, resnet
\end{IEEEkeywords}

\section{Introduction}
\label{sec:intro}

\begin{figure}[t]
\centering
\includegraphics[width=\linewidth]{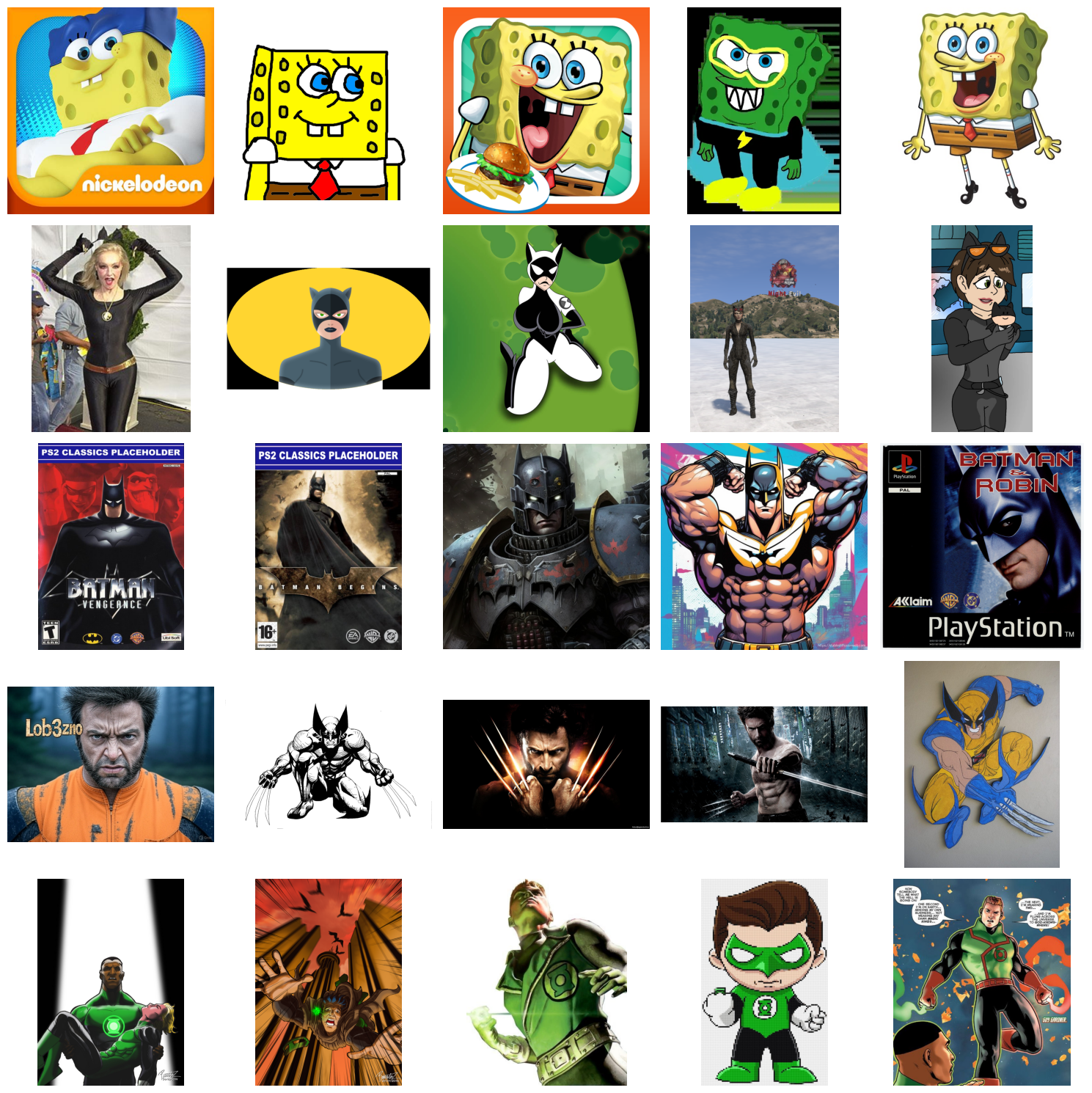}
\caption{Example copyrighted concepts and their five reference images used to build the reference database, showing the pose, lighting, and style variation the encoder must be invariant to.}
\label{fig:images}
\end{figure}

Recent advances in generative AI have made it relatively easy to create realistic digital content across modalities such as images, video, graphics, websites, art, and news media \cite {blattmann2023stable}. Within Computer Vision, AI-generated content has become one of the most important areas of research. Generative adversarial models initially gained popularity for their ability to generate photorealistic images. Following that, diffusion models\cite{blattmann2023stable} and autoregressive transformers have emerged as the most successful models for image generation. Based on the capability of diffusion models, several video generation models have been developed. These include text-to-video\cite{singer2022make}, image-to-video, video-to-video\cite{luo2022clip4clip}, video-editing, video generation\cite{wang2023modelscope} conditioned on image+text and many other video generation applications. With these impressive video generation capabilities, there are several legal and ethical challenges that we have to address. Recent research has shown that image and video generation models can recreate copyrighted content from the internet or other sources\cite{somepalli2023diffusion,pizzi2022self}. In this paper, we focus on conditioning-based copyright propagation, where a copyrighted reference image is supplied as input to an image-to-video (I2V) pipeline; training-based memorization, where a model reproduces content memorized from its training data, is outside the scope of this work. We introduce a dataset for copyright detection in AI-generated video. Along with that, we propose a contrastive retrieval-based method for copyright detection. The goal is to provide a baseline and open dataset to promote more research in this area, leading to fair use of video generation models.

\subsection{Background and Motivation}

Stability AI is a company that has emerged as the leading model provider for image and video generation using AI. Currently, Stability AI is facing several lawsuits in the U.S. and the UK for scraping images to train its generative AI model, Stable Diffusion. One of the landmark lawsuit by Getty images alleged that Stability AI has used 12 million of its photographs, captions, and metadata without permission\cite{somepalli2023diffusion}. As evidence, they have shown that earlier versions of Stable Diffusion sometimes replicated Getty watermarks. Another lawsuit by Disney, NBCUniversal, and Warner Bros. against Midjourney focuses on the generation of copyrighted material by AI models. The studios allege that the AI platform mass-infringed their intellectual property by training models on copyrighted works to let users generate unauthorized, high-quality images and videos of iconic characters\cite{cooper2025files,somepalli2023diffusion}. 

In this paper, we have created a dataset for copyright leakage detection in AI-generated video, with 50 copyrighted concepts, 5 reference images each, and 5 videos per concept from both image-conditioned and prompt-only generators. We also propose a contrastive retrieval method that detects leakage by matching video frame embeddings against a database of copyrighted image embeddings. This way, instead of just giving a yes/no answer, our framework can also point out which exact copyrighted image is being leaked. Both are released as an open baseline to support fairer video generation research.

\subsection{Limitations of Existing Copyright Detection Methods}

Existing research on copyright leakage detection focuses on image generation\cite{roy2025guardians}. While image generation has a significant risk of copyright infringement\cite{douze20212021, roy2025guardians}, video generation can often generate copyrighted characters as well as personal identities. This makes it necessary to focus on the detection of copyrighted content for video-based models. Existing research also lacks open data sets necessary for continued research\cite{roy2025guardians,cooper2025files,somepalli2023diffusion,carlini2023extracting}. We aim to establish an introductory benchmark and tool chain that can encourage more work in this area.

\subsection{Contributions}

The main contributions of this work are summarized as follows:

\begin{itemize}

\item We introduce a \textbf{new benchmark data set} for conditioning-based copyright leakage detection in AI-generated videos, consisting of 50 copyrighted concepts, reference images, and paired image-conditioned and prompt-only generated videos.

\item We provide a \textbf{simple, reproducible retrieval baseline} that formulates copyright leakage detection as a video-to-image retrieval problem by matching aggregated video frame embeddings against a database of copyrighted reference image embeddings. Rather than proposing a new architecture or loss, we deliberately build on standard components (a ResNet-18 encoder trained with SupCon and nearest-neighbor search) so that the benchmark is paired with a transparent and easily extensible pipeline. We also establish an \textbf{evaluation protocol} for copyright leakage detection using both retrieval and classification metrics.

\item Extensive experiments demonstrate the effectiveness of the proposed framework, achieving high retrieval performance and a \textbf{detection accuracy of 99.6\%} on the proposed benchmark. Together, this data set and the AI tool set can encourage more research in this area and lead to mitigation methods.

\end{itemize}

\section{Related Work}

\subsection{Video Generation Models}

Diffusion models have emerged as the dominant paradigm for state-of-the-art high-fidelity video generation. Building on image diffusion, Video Diffusion Models \cite{ho2022video} and Make-A-Video \cite{singer2022make} extended denoising generation to the temporal domain, and more recent latent video diffusion approaches such as Stable Video Diffusion \cite{blattmann2023stable} scaled this to high-resolution, image-conditioned generation. In parallel, powerful text-to-image (T2I) diffusion models such as SDXL \cite{podell2024sdxl} have become widely adopted as keyframe generators in video production pipelines, whereas dedicated video generation frameworks, including AnimateDiff \cite{guo2023animatediff}, ModelScopeT2V \cite{wang2023modelscope} (and its community fine-tuned derivative Zeroscope), and CogVideoX \cite{yang2025cogvideox}, support both text-to-video (T2V) and image-to-video (I2V) synthesis. A defining property of these models is that they are conditioned on external images or text and trained on large web-scraped corpora, through which copyrighted visual content can propagate into their outputs. Concretely, copyrighted visual content may enter a generated video through two distinct mechanisms: training-based memorization, in which representations learned from a web-scale corpus reappear in outputs, and conditioning-based propagation, in which the visual identity of a reference image or prompt is preserved through the generation pipeline. The latter mechanism, prevalent in image-to-video and key-frame-driven production workflows, remains relatively underexplored and is the focus of this work.

\subsection{Copyright, Memorization, and Replication in Visual Generative Models}

The question of whether visual generative models reproduce copyrighted content has moved from a legal concern to an active research problem \cite{jia2025imitation, cooper2025files}. A growing body of work shows that large-scale diffusion models do not merely generalize but can memorize and reproduce individual training examples. Notably, Carlini et al. \cite{carlini2023extracting} demonstrated that near-verbatim copies of training images can be extracted from diffusion models, while Somepalli et al. \cite{somepalli2023diffusion} showed that a non-trivial fraction of generated images are partial or full replications of specific training samples, detectable by retrieving the nearest training image for each generation. Follow-up work analyzed the causes of this behavior and proposed mitigations at generation time \cite{somepalli2023understanding, wen2024detecting}. A complementary, mitigation-oriented line removes protected visual concepts or styles from the model itself through concept erasure \cite{gandikota2023erasing} or concept ablation \cite{kumari2023ablating}. However, existing approaches almost exclusively address the training-based replication channel, operating on static images, matching generated images against training images in a single modality, and typically assuming access to the model or its training data. In contrast, the conditioning-based propagation channel, where a copyrighted reference is deliberately supplied as input to an I2V pipeline, and the video setting more broadly remain relatively underexplored. Yet this setting is a growing and legally salient risk: modern production workflows routinely animate reference frames into full sequences, and the resulting videos may reproduce protected characters, costumes, or cinematic identities without authorization. Crucially, whether such propagation is legally infringing is a downstream determination; the technical task we address is to flag and attribute candidate reproductions to a reference database, providing an interpretable audit signal rather than a legal verdict

\subsection{Contrastive Representation Learning}

Contrastive representation learning has emerged as a dominant paradigm for self-supervised feature extraction, mapping data into a latent embedding space where positive pairs are pulled together while negative pairs are pushed apart, yielding representations invariant to nuisance transformations such as cropping, color jitter, and blur\cite{chen2020simple, he2020momentum, oord2018representation, xu2021videoclip}. Supervised contrastive learning (SupCon) \cite{khosla2020supervised, chowdhury2025supervised} generalizes this objective to leverage label information, pulling together all samples sharing a class while separating differing classes, which produces tighter, more cluster-structured embeddings. Beyond hand-crafted augmentations, self-distillation methods such as DINO \cite{oquab2023dinov2} learn similarly discriminative features without negatives, and large vision-language models such as CLIP \cite{radford2021learning} provide strong general-purpose embeddings from image–text contrast. Overall, the invariance and instance-discriminative structure induced by these contrastive objectives are precisely the properties needed to recognize the same copyrighted identity across pose, lighting, and stylistic variations introduced by generative pipelines, making contrastive representations a natural foundation for a similarity-based leakage detector.

\subsection{Cross-Modal and Video Retrieval}

Detecting copyrighted content through similarity search is closely related to content-based copy detection, which seeks to identify near-duplicate or transformed versions of a reference item. In the image domain, the DISC21 benchmark \cite{douze20212021} spurred descriptors robust to strong edits, most notably SSCD \cite{pizzi2022self}, a self-supervised copy-detection embedding that treats copy detection as instance-level retrieval. In the video domain, segment- and video-level copy detection has been studied on large benchmarks such as VCSL \cite{he2022large}, and cross-modal retrieval methods built on contrastive vision–language embeddings \cite{radford2021learning, luo2022clip4clip} match content across text, image, and video. However, in generated videos, frames may deviate from any real reference through synthesis artifacts, motion, and stylistic drift. Thus, matching it against static, real copyrighted images is an inherently asymmetric and cross-domain problem. Moreover, conditioning-based propagation preserves identity (character, costume, palette) rather than pixels, requiring identity or visual concept-level rather than instance-level matching. We address both by aggregating per-frame embeddings into a single video-level representation and retrieving against a concept-structured reference database.

To the best of our knowledge, there is no open-source dataset available for research on copyright leakage and potentially illegal video generation. No existing framework performs retrospective, model-agnostic detection and tracing of copyrighted visual content in generated video via video-to-image retrieval. This gap between the growing risk of copyright leakage and the lack of dedicated tools directly motivates the benchmark and framework proposed in this paper.

\begin{figure}[t]
\centering
\includegraphics[width=\linewidth]{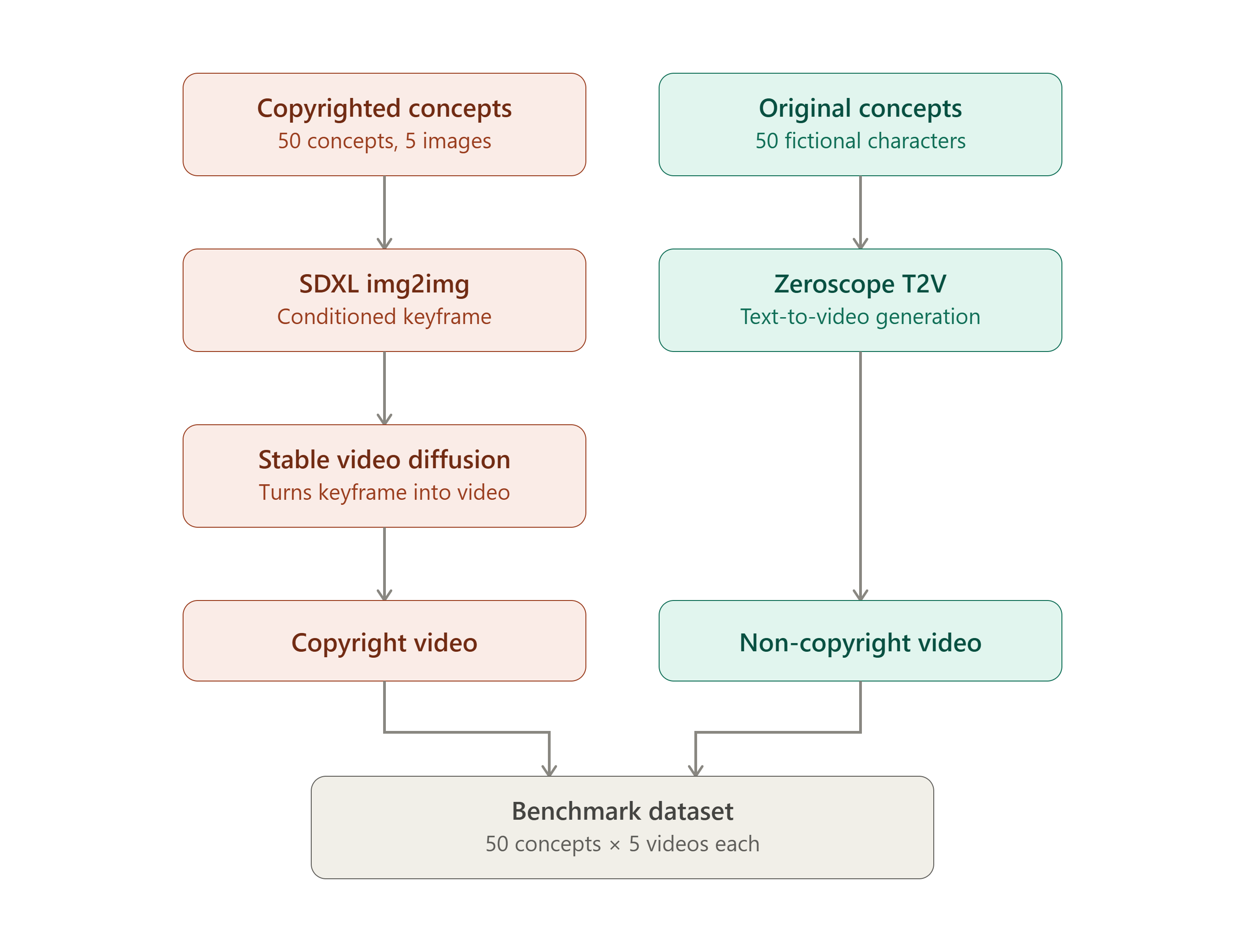}
\caption{Construction of the benchmark data set. Reference images from 50 copyrighted concepts are animated into copyright videos via image-conditioned generation (SDXL img2img followed by Stable Video Diffusion), while 50 original fictional concepts are rendered directly from text prompts using Zeroscope to form the non-copyright videos. The two branches together form the final benchmark.}
\label{fig:data}
\end{figure}

\section{Method}

\begin{figure*}[t]
\centering
\includegraphics[width=\linewidth]{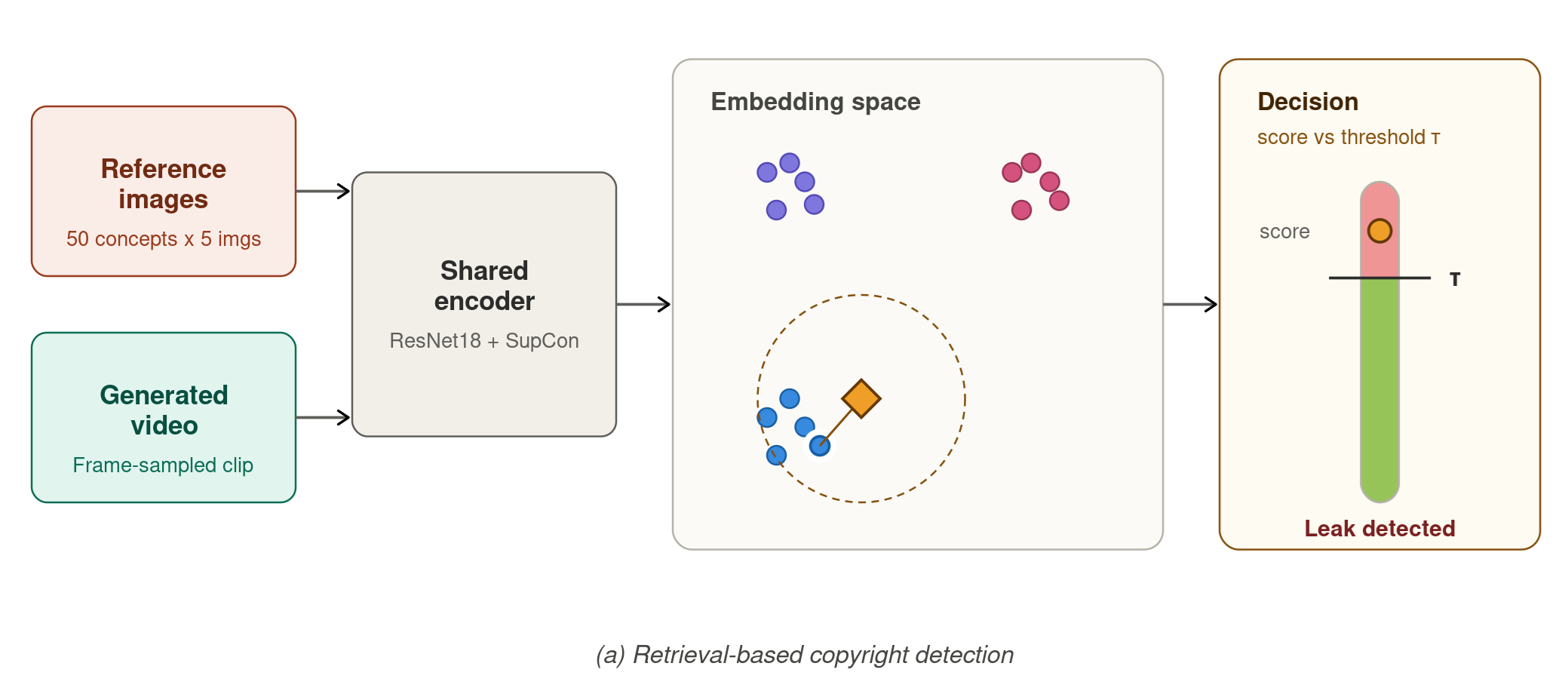}
\caption{Overview of the proposed retrieval-based copyright detection framework. Reference images and generated video frames are mapped into a shared embedding space by a contrastive encoder (ResNet-18 trained with SupCon). The resulting video embedding is compared against the reference database, and a copyright leak is declared when its similarity to the nearest reference exceeds a threshold $\tau$.}
\label{fig:method}
\end{figure*}

\subsection{Problem Formulation}

Let $\mathcal{C}=\{c_1,\dots,c_K\}$ denote the set of K copyrighted concepts, each associated with a set of images $R_k=\{I_k^1,\dots,I_k^5\}$. Here, $c_{i}$ is a concept and $I_{i}$ is an image. Given a generated video $V=(F_1,\dots,F_T)$, the goal is to check whether it reproduces any of the reference images. Our method tries to learn an encoder $f_\theta:\mathbb{R}^{H\times W\times 3}\rightarrow\mathbb{R}^d$ that maps an image to an embedding $z_{i}=f_\theta(I_{i})$. We build an image embedding database $\mathcal{D}=\{(z_k^j,c_k)\}$ built from $R_{K}$. For a generated video $V$, we compute a single video embedding $v$ and compare it against our image database $D$. Copyright leakage is then cast as a nearest-neighbor decision: $V$ is flagged as leaking concept $c_k$ if the similarity between $v$ and its closest entry in $\mathcal{D}$ exceeds a threshold $\tau$.

\subsection{Benchmark Dataset}

The benchmark is developed using $K{=}50$ copyrighted concepts, each with 5 images collected from \emph{archive.org}, resulting in a total dataset of 250 images. For each concept, we generate 5 videos using the video generation model with a prompt and an image as input. These 250 videos become positive samples for copyright leakage videos. We also create 250 videos based on 50 different concepts using only the text prompt as input to the video generation model. These become the negative samples for copyright leakage.

\begin{figure*}[t]
\centering
\includegraphics[width=\linewidth]{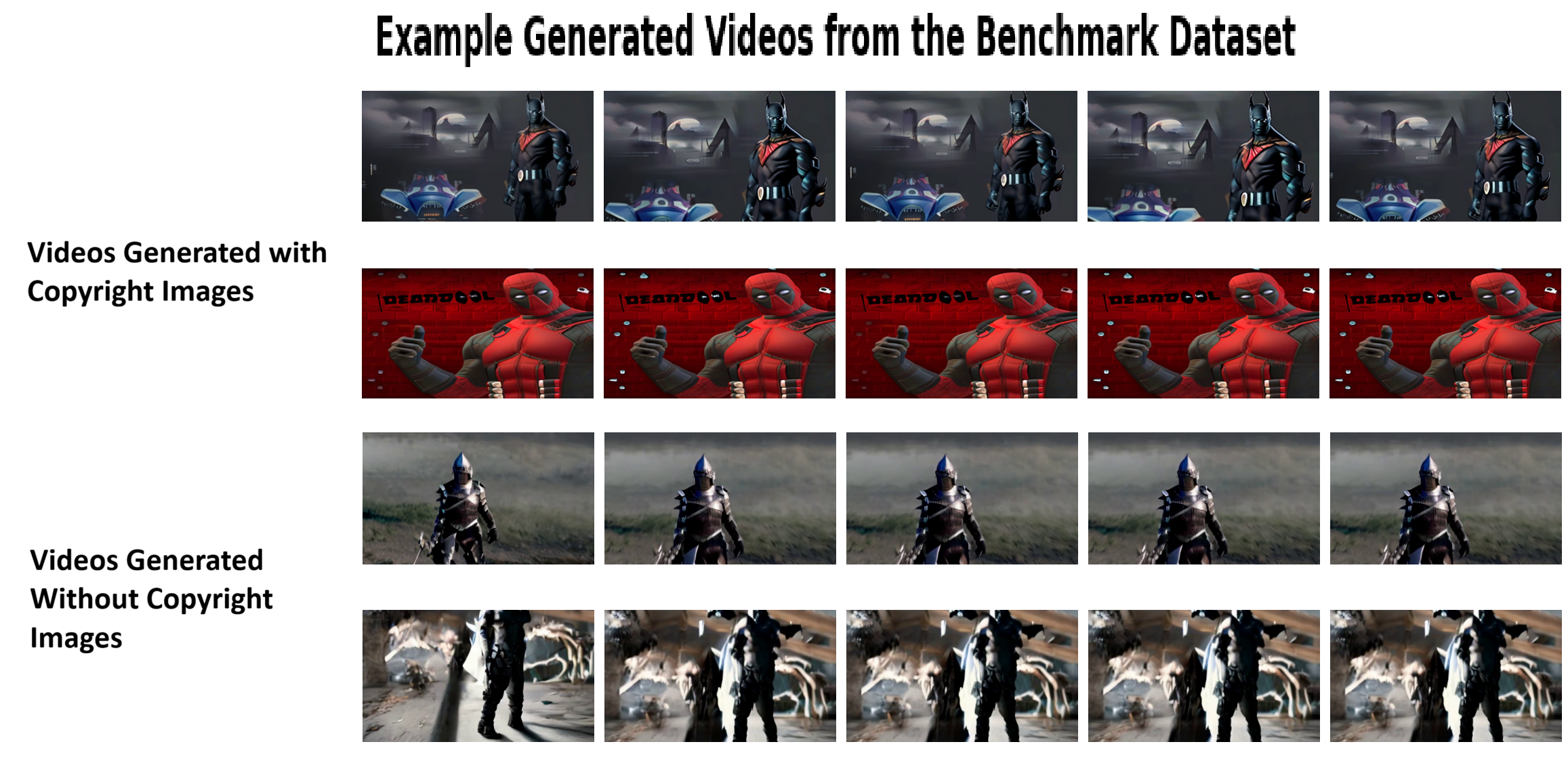}
\caption{Example frames from the benchmark's generated videos. Top two rows: image-conditioned videos built from copyrighted reference images. Bottom two rows: prompt-only videos generated without any copyrighted reference. For both generation settings, the character-name-conditioned prompt used was: ``Generate a cinematic video closely recreating the character. Preserve facial structure, costume, lighting, and mood. Subtle natural motion only. A [concept] character.''}
\label{fig:examples}
\end{figure*}

\subsubsection{Copyright Video Generation}

For each reference image $I_k^j$, we synthesize a keyframe using an image-conditioned diffusion model $G_{\text{img}}$, $\hat{I}=G_{\text{img}}(I_k^j,p;s)$, where $p$ is a fixed cinematic prompt and $s$ is the denoising strength controlling how closely $\hat{I}$ preserves the reference identity. The keyframe is then used as input to a video diffusion model generating a video, $V=G_{\text{svd}}(\hat{I})$. This two-stage process yields videos that are conditioned on copyrighted visual content.

\subsubsection{Non-Copyright Video Generation}

For the negative set, videos are produced by a text-to-video model $G_{\text{t2v}}$ using a concept name and a generic prompt, $V=G_{\text{t2v}}(p_k)$, with no image conditioning. Because $G_{\text{t2v}}$ never observes a reference image, these videos share the same generation pipeline structure as the copyright set but contain no copyrighted visual signal. 

\subsection{Contrastive Image Representation Learning}

We have used contrastive learning to generate embeddings from the images and videos. Given an input anchor image, contrastive learning generates an augmented view and learns representations by maximizing agreement between the anchor and positive sample, while increasing the distance between the anchor and negative sample in the feature space. Contrastive learning has an encoder that can be used to generate representations from images and videos. The encoder backbone is a ResNet-18 followed by a projection head $g_\phi$, producing $z=g_\phi(f_\theta(I))/\lVert g_\phi(f_\theta(I))\rVert$. We chose ResNet-18 deliberately: with only 250 reference images, a compact backbone reduces the risk of overfitting during contrastive training and keeps frame-level screening computationally inexpensive. Larger foundation backbones such as CLIP \cite{radford2021learning} or DINOv2 \cite{oquab2023dinov2} are natural drop-in replacements that we leave for future work. For each image, we create two augmented views and train with the supervised contrastive loss: 
$\mathcal{L}_{\text{SupCon}} = -\frac{1}{N}\sum_{i=1}^{N}\frac{1}{|P(i)|}\sum_{p\in P(i)}\log\frac{\exp(z_i\cdot z_p/\tau)}{\sum_{a\in A(i)}\exp(z_i\cdot z_a/\tau)}$

where $P(i)$ is the set of augmented views of anchor image $i$, $A(i)$ is all other views in the batch, and $\tau$ is a temperature parameter. This objective pulls embeddings of the same concept together and pushes different concepts apart.

\subsection{Reference Database Construction}

After training, we use the encoder to generate features from images $z=f_\theta(I)/\lVert f_\theta(I)\rVert$, forming the database $\mathcal{D}=\{(z_k^j,c_k)\}_{k=1,j=1}^{K,5}$. $\mathcal{D}$ is indexed for efficient nearest-neighbor search and serves as the fixed target when searching for copyright leakage among videos.

\subsection{Video Representation}

We extract frames from every video $\{F_t\}_{t=1}^{T}$, and every frame is passed through the same encoder $f_\theta$ to obtain frame embeddings $\{u_t\}$. We compute the average of these features to create a single video-level embedding.
\begin{equation}
v = \frac{1}{T}\sum_{t=1}^{T} u_t \Big/ \Big\lVert \frac{1}{T}\sum_{t=1}^{T} u_t \Big\rVert ,
\end{equation}
This feature summarizes the video's overall visual identity while remaining directly comparable to reference embeddings in $\mathcal{D}$. Retrieval is therefore performed in a shared image embedding space (video-to-image retrieval) rather than through full spatio-temporal video modeling. Frame averaging suits the conditioning-based setting, where the copyrighted identity (character, costume, palette) is carried by appearance and persists across frames. However, it discards temporal order, so infringement expressed only through motion or sequence (e.g., choreography or a specific animated sequence) is beyond the scope of this work.

\subsection{Copyright Leakage Detection}

Given a video embedding $v$, we compute its similarity to every entry in the image feature database $\mathcal{D}$ as $s(v,z)=v^\top z$ and retrieve the nearest reference $z^\star=\arg\max_{z\in\mathcal{D}} s(v,z)$. Using that image, we also fetch the nearest concept label $c^\star$. The video is declared a copyright leak of $c^\star$ if $s(v,z^\star)\geq\tau$, and clean otherwise. So that $\tau$ becomes the parameter that directly controls the sensitivity of detection. Because the decision is based on nearest neighbor search, the framework also returns the specific copyrighted concept, providing retrieved reference evidence for each decision rather than an opaque label. This output is a screening signal for human or legal review, not a determination of infringement or fair use, which depends on legal factors such as transformative use that a similarity score cannot capture.

\begin{figure}[t]
\centering
\includegraphics[width=\linewidth]{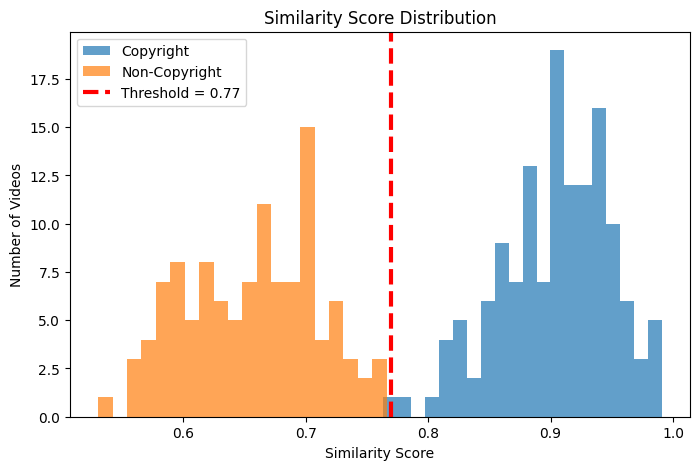}
\caption{Distribution of similarity scores between generated videos and their nearest reference embedding, split by copyright and non-copyright videos, with the learned threshold $\tau=0.77$ (red dashed line) separating the two populations.}
\label{fig:sim}
\end{figure}

Fig.~\ref{fig:data} summarizes how the benchmark's copyright and non-copyright videos are generated from their respective concept sets, and Fig.~\ref{fig:method} illustrates the full detection pipeline, from contrastive encoding to the threshold-based copyright decision. The use of same encoder allows threshold rule to generalize across both video generation pipelines shown in Fig.~\ref{fig:data}.

\begin{table*}[t]
\centering
\caption{Ablation of the decision strategy: the proposed retrieval-based thresholding versus standard classifiers using identical ResNet-18 + SupCon embeddings. All methods use the same encoder, reference database, and train/test split, so this is not a comparison with state-of-the-art copy detection methods. Best results are shown in bold.}
\label{tab:classifier_baselines}

\begin{tabular}{lccccc}
\hline
Method & Accuracy & Precision & Recall & F1 & ROC-AUC\\
\hline
\textbf{Proposed: Retrieval + Threshold} & \textbf{0.9960} & \textbf{1.0000} & \textbf{0.9928} & \textbf{0.9964} & \textbf{0.9998}\\
Linear SVM & 0.8924 & 0.9308 & 0.8705 & 0.8996 & 0.9777\\
Random Forest & 0.8765 & 0.8649 & 0.9209 & 0.8920 & 0.9442\\
Logistic Regression & 0.8645 & 0.8723 & 0.8849 & 0.8786 & 0.9392\\
k-NN ($k=5$) & 0.8327 & 0.9802 & 0.7122 & 0.8250 & 0.9472\\
\hline
\end{tabular}
\end{table*}

\section{Results}

\subsection{Dataset}

The benchmark comprises 50 copyrighted concepts and 50 original non-copyright concepts, each with 5 generated videos, giving 500 videos in total across the image-conditioned and prompt-only pipelines. Fig.~\ref{fig:images} shows example concepts with their five reference images. Videos are decoded frame-by-frame, resized to $224\times224$, and normalized before they are passed to the ResNet-18 encoder trained with contrastive learning. Each video is represented by the average of features and compared against the reference database via cosine similarity. 40\% of the videos were used to create the decision threshold, and the rest were used for testing and analysis. We report binary classification accuracy, precision, recall, and confusion metrics computed on the held-out test set using the threshold learned on the training split.

\begin{figure}[!t]
\centering
\includegraphics[width=0.9\linewidth]{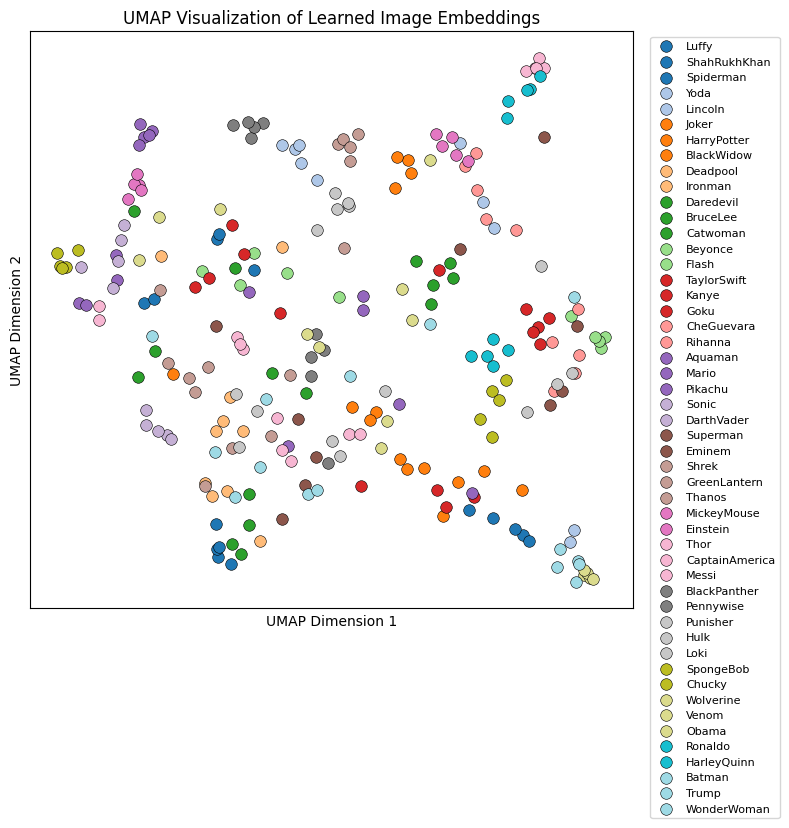}
\caption{UMAP projection of the learned reference image embeddings, colored by concept. Images of the same copyrighted character form tight, well-separated clusters.}
\label{fig:imgemb}
\end{figure}

\subsection{Copyright Detection Results}

Using a threshold of $\tau=0.77$ selected on the training split, our method achieves 99.6\% accuracy on the test set, with 138 of 139 copyright videos correctly flagged and all 112 non-copyright videos correctly cleared. This corresponds to a precision of 100\% and a recall of 99.3\%, with only a single false negative and zero false positives. Fig.~\ref{fig:sim} shows that the similarity scores of copyright and non-copyright videos form two well-separated distributions, with the learned threshold falling cleanly in the gap between them.

\begin{figure}[!t]
\centering
\includegraphics[width=0.9\linewidth]{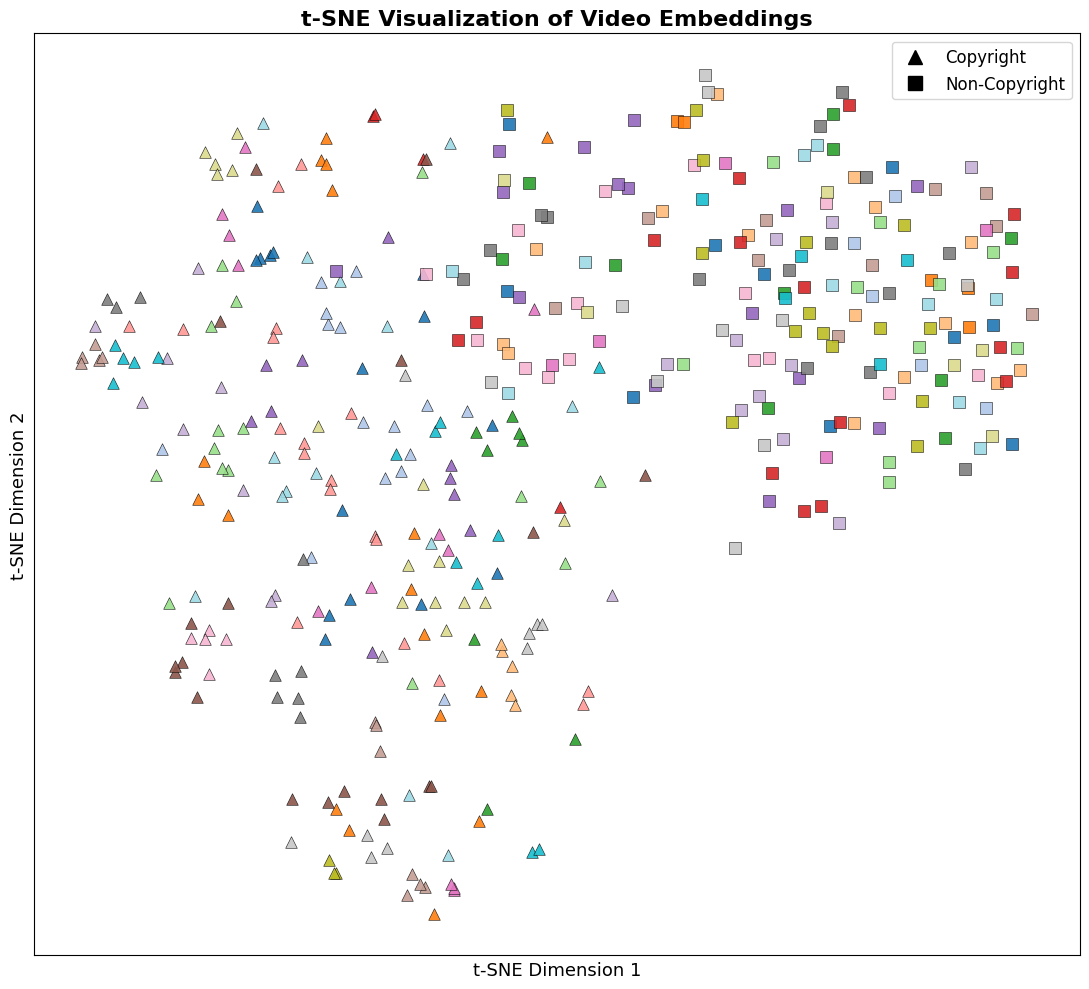}
\caption{t-SNE projection of generated video embeddings. Copyright videos (triangles) and non-copyright videos (squares) occupy largely distinct regions of the space.}
\label{fig:videmb}
\end{figure}

\subsection{Retrieval Results}

Beyond the binary decision, our framework returns the specific reference image responsible for the highest similarity score, allowing each detected copyright leak to be traced back to a source image. In practice, the top-ranked retrieval consistently corresponds to the correct ground-truth concept for copyright videos, confirming that the embedding space is discriminative enough for localization and not just detection. Fig.~\ref{fig:imgemb} visualizes the learned reference image embeddings with UMAP, showing that images of the same copyrighted concept are grouped together. Fig.~\ref{fig:videmb} shows the corresponding t-SNE projection of generated video embeddings, where copyright and non-copyright are clustered in different regions with few overlaps.

\subsection{Qualitative Results}

Fig.~\ref{fig:examples} shows some example frames from the generated videos. The image-conditioned videos (top two rows) visibly preserve the identity, costume, and color palette of their copyrighted reference, while the prompt-only videos (bottom two rows) depict generic characters with no identifiable source. This contrast shows the difference in similarity scores between copyright and non-copyright videos.

\subsection{Comparison with Standard Classifiers}

To compare against standard classifiers, we used the same ResNet-18 + SupCon encoder, reference database, and train/test split as the proposed method. Each video was represented by a cosine similarity vector computed between its embedding and all reference image embeddings (250 dimensions for the copyright-only database and 440 dimensions after adding non-copyright references). Logistic Regression, Linear SVM, Random Forest, and feature-space k-NN were trained using these similarity vectors, while the proposed retrieval method and image-database k-NN directly queried the reference database at inference time. Thus, all methods used identical features and differed only in their decision strategy; this experiment isolates the effect of the decision rule rather than comparing against state-of-the-art video copy detection descriptors such as SSCD \cite{pizzi2022self} or frozen foundation-model embeddings such as CLIP \cite{radford2021learning} and DINOv2 \cite{oquab2023dinov2}, which we leave for future work. Table~\ref{tab:classifier_baselines} shows that the proposed retrieval-based thresholding approach consistently outperformed all competing methods.

\subsection{Ablation Study, Limitations and Future Work}

In this paper, we use two different strategies to analyze video embeddings: feature averaging across all frames and taking the maximum per-frame similarity to the database. The late fusion of feature averaging allows a smoother, more stable similarity score, while max-frame similarity allows it to be more interpretable.

Beyond the detection framework and experiments, the value of this benchmark lies in its paired construction: because every copyrighted concept has a matched non-copyright counterpart generated with the same pipeline, it provides a controlled comparison for examining image conditioning. However, the benchmark should be viewed as a controlled pilot study: with 50 concepts, 250 reference images, and 500 videos, its scale is small, and results may not transfer to in-the-wild user-generated videos, to subtler forms of copying such as style or likeness imitation, or to videos from unknown generators. The 99.6\% accuracy may also be influenced by the controlled data set construction, and frame averaging does not capture infringement expressed only through motion. In addition, we have not yet evaluated robustness to common evasion edits such as cropping, color jittering, additive noise, or inpainting, which a deployed copyright detector must withstand. Our method also relies on a single embedding model, and evaluating alternative embeddings is an important direction for future work. A further deployment consideration is the privacy of user-provided reference images and audited videos. Extending copyright leakage detectors to a federated setting \cite{mcmahan2017communication, tonmoy2025leveraging, chowdhury2024acfed, zhang2021survey} is therefore a promising direction for privacy-preserving auditing across content platforms.  In addition, future work could investigate lightweight and compression techniques \cite{tonmoy2025attention,rokh2023comprehensive,tonmoy2024lightweight,chowdhury2024acfed} to reduce the computational cost of frame-level copyright screening frameworks, thereby enabling on-device decision-making and advancing edge intelligence in AI-based copyright-auditing systems.

We release the data set, trained models, source code, reference embeddings, and evaluation protocol openly, in the hope of encouraging further research into detection and, ultimately, mitigation strategies that allow video generation models to be developed and deployed more fairly.

\section{Conclusion}

In this paper, we address a largely unexplored problem: detecting conditioning-based copyright leakage in AI-generated video, where a copyrighted reference image is propagated through an image-to-video pipeline. As video generation models become stronger, they are increasingly capable of creating realistic video conditioned on copyrighted images and prompts. The models are able to reproduce protected characters, artwork, real human identities, etc. This creates legal and ethical concerns for model developers, platforms, and rights holders alike. To address this gap, we introduced a new benchmark data set of 50 copyrighted concepts, each paired with reference images and videos. We also create a set of non-copyright videos generated through an identical pipeline. This paired construction helps examine the role of image conditioning in copyright leakage and enables precise, reproducible evaluation.

Beyond copyright, this paper connects to the deepfake video problem, where generated videos are misused for impersonation and misinformation. Our retrieval approach can be extended for identity verification and content provenance too. We release our complete tool set, including the data set, models, embeddings, source code, and evaluation protocol, publicly. Future work includes extending detection to training-based memorization, where text-to-video models reproduce copyrighted content without image conditioning, exploring mitigation such as concept erasure, and testing robustness against adversarial evasion. We hope this encourages more research towards safer video generation. The data set will be publicly released upon publication; the data set link is omitted from the current submission to preserve anonymity.



\bibliographystyle{ieeetr}
\bibliography{ref}

\end{document}